\documentclass[runningheads]{llncs}
\usepackage{graphicx}

\usepackage{tikz}
\usepackage{amsmath,amssymb} 
\usepackage{color}

\usepackage[breaklinks,colorlinks]{hyperref}

\usepackage{multirow}
\usepackage{subcaption}
\usepackage{soul}
\usepackage[normalem]{ulem}
\usepackage{tabularx, booktabs}
\newcolumntype{L}[1]{>{\raggedright\let\newline\\\arraybackslash\hspace{0pt}}m{#1}}
\newcolumntype{C}[1]{>{\centering\let\newline\\\arraybackslash\hspace{0pt}}m{#1}}
\newcolumntype{R}[1]{>{\raggedleft\let\newline\\\arraybackslash\hspace{0pt}}m{#1}}

\newif\ifdraft
\draftfalse
\drafttrue

\usepackage{xcolor}
\definecolor{orange}{rgb}{1,0.5,0}
\definecolor{violet}{RGB}{70,0,170}

\ifdraft
 \newcommand{\PF}[1]{{\color{red}{\bf PF: #1}}}
 
 \newcommand{\MS}[1]{{\color{green}{\bf MS: #1}}}
 
 \newcommand{\ZD}[1]{{\color{violet}{\bf ZD: #1}}}
 
 \newcommand{\YH}[1]{{\color{blue}{\bf YH: #1}}}
 
 \newcommand{\SPE}[1]{{\color{orange}{\bf SS: #1}}}
 
 \newcommand{\WJ}[1]{{\color{orange}{\bf WJ: #1}}}
 
\else
 \newcommand{\PF}[1]{}
 
 \newcommand{\KY}[1]{}
 
 \newcommand{\MS}[1]{}
 
 \newcommand{\ZD}[1]{}
 
 \newcommand{\YH}[1]{}
 
 \newcommand{\SPE}[1]{}
 
 \newcommand{\WJ}[1]{}
 
\fi

\newcommand{\comment}[1]{}

\newcommand{\bI}{\mathbf{I}}
\newcommand{\bF}{\mathbf{F}}
\newcommand{\bK}{\mathbf{K}}

\newcommand{\bR}{\mathbf{R}}

\newcommand{\bu}{\mathbf{u}}

\newcommand{\f}{\mathbf{f}}

\begin{document}
\pagestyle{headings}
\mainmatter

\title{Perspective Flow Aggregation for \\ Data-Limited 6D Object Pose Estimation}


%

\author{%
	{Yinlin Hu $^{1,2}$, \quad Pascal Fua $^1$, \quad Mathieu Salzmann $^{1,2}$}
}
\titlerunning{PFA-Pose}
\authorrunning{Hu et al.}
\institute{
    {\small $^1$ EPFL CVLab, \quad $^2$ ClearSpace SA} \\
    \email{{\tt \small huyinlin@gmail.com,\quad\{firstname.lastname\}@epfl.ch}}
}

\maketitle


\vspace{-1em} 
\begin{abstract}

Most recent 6D object pose estimation methods, including unsupervised ones, require many real training images. Unfortunately, for some applications, such as those in space or deep under water, acquiring real images, even unannotated, is virtually impossible. In this paper, we propose a method that can be trained solely on synthetic images, or optionally using a few  additional real ones. Given a rough pose estimate obtained from a first network, it uses a second network to predict a dense 2D correspondence field between the image rendered using the rough pose and the real image and infers the required pose correction. This approach is much less sensitive to the domain shift between synthetic and real images than state-of-the-art methods. It performs on par with methods that require annotated real images for training when not using any, and outperforms them considerably when using as few as twenty real images. 


\end{abstract}




\section{Introduction}
\label{sec:introduction}

\begin{figure}[t]
    \begin{center}
    %
    \addtolength{\tabcolsep}{-5pt}  
    \begin{tabular}{c@{\hskip 0.2em}c@{\hskip 0.2em}c@{\hskip 0.2em}c@{\hskip 0.2em}c}
     \includegraphics[width=0.19\linewidth, trim=40 0 120 0, clip]{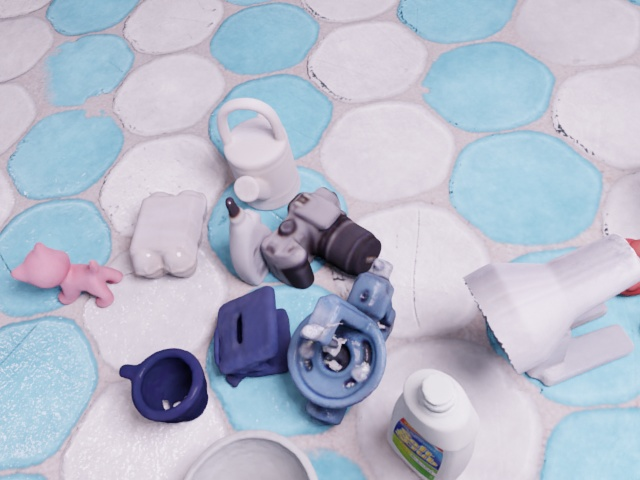} &
     \includegraphics[width=0.19\linewidth]{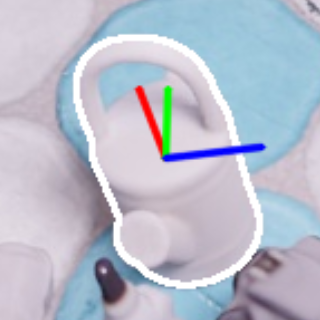} &
     \includegraphics[width=0.19\linewidth]{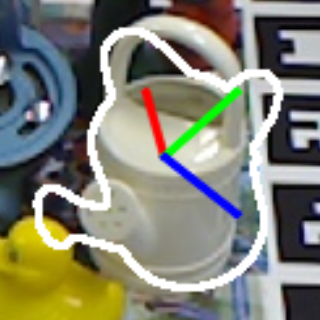} &
     \includegraphics[width=0.19\linewidth]{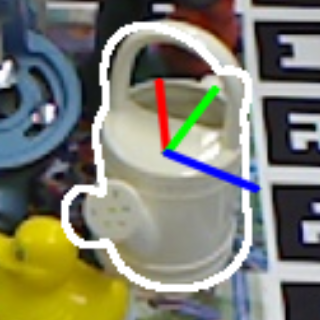} &
     \includegraphics[width=0.19\linewidth]{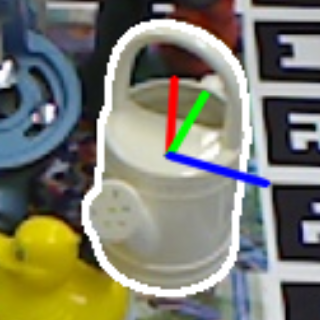} \\
    {\small (a)} & {\small (b)} & {\small (c)} & {\small (d)} & {\small (e)} \\

    \end{tabular}
    \addtolength{\tabcolsep}{5pt}  
    \end{center}
    \vspace{-6mm}
    \caption{{\bf Data-limited 6D object pose estimation.} {\bf (a)} In the absence of real data, one can train a model using synthesized images~\cite{hodan19a}. {\bf (b)} Although the resulting accuracy on synthetic data is great, {\bf (c)} that on real images is significantly worse. {\bf (d)} While the common iterative pose refinement approach can help, it still suffers from the synthetic-to-real domain gap~\cite{Li18a}.
    {\bf (e)} Our non-iterative strategy generalizes much better to real images despite being trained only on synthetic data.
    }
    \label{fig:teaser}
\end{figure}

Estimating the 6D pose of a target object is at the heart of many robotics, quality control, augmented reality applications, among others. When ample amounts of annotated real images are available, deep learning-based methods now deliver excellent results~\cite{Hodan18,Peng19a,Park19a,Wang19d,Song20a}. Otherwise, the most common approach is to use synthetic data instead~\cite{Li18a,Xiang18b,Hu21a}. However, even when sophisticated domain adaptations techniques are used to bridge the domain gap between the synthetic and real data~\cite{Rad18,Li19a,Hu21a}, the results are still noticeably worse than when training with annotated real images, as illustrated by Fig.~\ref{fig:teaser}.
 
\begin{figure*}[t]
    \begin{center}
    %
    \addtolength{\tabcolsep}{-5pt}  
    \begin{tabular}{cc}
    \includegraphics[width=0.45\linewidth]{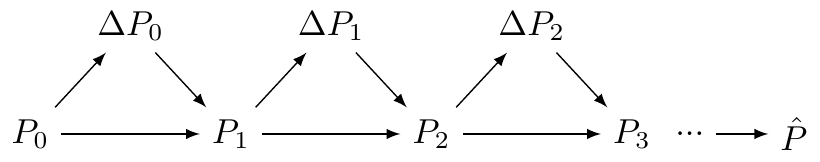} &
    \hspace{3em}\includegraphics[width=0.45\linewidth]{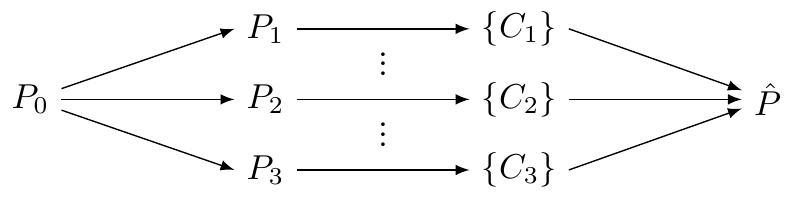} \\
    {\small (a) The iterative strategy} &
    \hspace{3em}{\small (b) Our non-iterative strategy} \\
    \end{tabular}
    \addtolength{\tabcolsep}{5pt}  
    \end{center}
    \vspace{-6mm}
    \caption{{\bf Different pose refinement paradigms.} {\bf (a)} Given an initial pose $P_0$, existing refinement strategies estimate a pose difference $\Delta P_0$ from the input image and the image rendered according to $P_0$, generating a new intermediate pose $P_1$. They then iterate this process until it converges to the final pose $\hat P$~\cite{Li18a,Yann20a}. This strategy relies on estimating a delta \emph{pose} from the input images by extracting global object features. These features contain high-level information, and we observed them not to generalize well across domains. {\bf (b)} By contrast, our strategy queries a set of discrete poses \{$P_1$, $P_2$, $P_3$, \dots\} that are near the initial pose $P_0$ from pre-rendered exemplars, and computes the final pose $\hat P$ in one shot by combining all the correspondences \{$C_i$\} established between the exemplars and the input. Estimating \emph{dense 2D-to-2D local correspondences} forces the supervision of our training to occur at the pixel-level, not at the image-level as in {\bf (a)}. This makes our DNN learn to extract features that contain lower-level information and thus generalize across domains. In principle, our method can easily be extended into an iterative strategy, using the refined pose as a new initial one. However, we found a single iteration to already be sufficiently accurate.}
    \label{fig:refine_compare}
\end{figure*}

Pose refinement offers an effective solution to this problem: An auxiliary network learns to correct the mistakes made by the network trained on synthetic data when fed with real data~\cite{Rad17,Li18a,Zakharov19a,Yann20a}. 
The most common refinement strategy is to render the object using the current pose estimate, predict the 6D difference with an auxiliary network taking as input the rendered image and the input one, and correct the estimate accordingly. As illustrated by Fig.~\ref{fig:refine_compare}(a), this process is performed iteratively. Not only does this involve a potentially expensive rendering at each iteration,
but it also is sensitive to object occlusions and background clutter, which cannot be modeled in the rendering step.
Even more problematically, most of these methods still require numerous real images for training purposes, and there are applications for which such images are simply not available.  For example, for 6D pose estimation in space~\cite{Kisantal20,Hu21a} or deep under water~\cite{Joshi20a,Risholm21a}, no real images of the target object may be available, only a CAD model and conjectures about what it now looks like after decades in a harsh environment. These are the scenarios we will refer to as {\it data-limited}.


To overcome these problems, we introduce the non-iterative pose refinement strategy depicted by Fig.~\ref{fig:refine_compare}(b).
We again start from a rough initial pose but, instead of predicting a delta pose, we estimate a dense 2D correspondence field between the image rendered with the initial pose and the input one. We then use these correspondences to compute the 6D correction algebraically. Our approach is simple but motivated by the observation that predicting dense 2D-to-2D matches is much more robust to the synthetic-to-real domain gap than predicting a pose difference directly from the image pair, as shown in Fig.~\ref{fig:teaser}(d-e). Furthermore, this strategy naturally handles the object occlusions and is less sensitive to the background clutter.

Furthermore, instead of synthesizing images given the rough initial pose, which requires on-the-fly rendering, we find nearest neighbors among pre-rendered exemplars and estimate the dense 2D correspondences between these neighbors and the real input. This serves several purposes. First, it makes the computation much faster. Second, it makes the final accuracy less dependent on the quality of the initial pose, which only serves as a query for exemplars. Third, as multiple exemplars are independent of each other, we can process them simultaneously. Finally, multiple exemplars deliver complementary perspectives about the real input, which we combine for increased robustness.

We evaluate our pose refinement framework on the challenging Occluded-LINEMOD~\cite{Krull15} and YCB-V~\cite{Xiang18b}  datasets, and demonstrate that it is significantly more efficient and accurate than iterative frameworks. It performs on par with state-of-the-art methods that require annotated real images for training when not using any, and outperforms them considerably when using as few as twenty real images.


\section{Related Work}
\label{sec:related}

{\bf 6D pose estimation} is currently dominated by neural network-based methods~\cite{Hu19a,Peng19a,Song20a,Sock20,Hu21a,Di21a}. However, most of their designs are still consistent with traditional techniques. That is, they first establish 3D-to-2D correspondences~\cite{Lowe04,Tola10,Trzcinski12c} and then use a Perspective-n-Points (PnP) algorithm~\cite{Lepetit09,Zheng13,Kneip14,Ferraz14}. In practice, these correspondences are obtained by predicting either the 2D locations of predefined 3D keypoints~\cite{Kehl17,Rad17,Tekin18a,Oberweger18,Jafari18}, or the 3D positions of the pixels within the object mask~\cite{Zakharov19a,Li19a,Hodan20}. These methods have been shown to outperform those that directly regress the 6D pose~\cite{Xiang18b}, which are potentially sensitive to object occlusions. Nevertheless, most of these methods require large amounts of annotated real training data to yield accurate predictions. Here, we propose a pose refinement strategy that allows us to produce accurate pose estimates using only synthetic training data.

{\bf 6D pose refinement}~\cite{Kehl17,Rad17,Zakharov19a,Sundermeyer20a} aims to improve an initial rough pose estimate, obtained, for example, by a network trained only on synthetic data.
In this context, DeepIM~\cite{Li18a} and CosyPose~\cite{Yann20a} iteratively render the object in the current pose estimate and predict the 6D pose difference between the rendered and input images. However, learning to predict a pose difference directly does not easily generalize to different domains, and these methods thus also require annotated data. Furthermore, 
the on-the-fly rendering performed at each iteration makes these algorithms computationally demanding. Finally, these methods are sensitive to object occlusions and background clutter, which cannot be modeled in the rendering process. Here, instead, we propose a non-iterative method based on dense 2D correspondences. Thanks to our use of offline-generated exemplars and its non-iterative nature, it is much more efficient than existing refinement methods. Furthermore, it inherently handles occlusions and clutter, and, as we will demonstrate empirically, generalizes easily to new domains.
The method in~\cite{Manhardt18} also uses 2D information for pose refinement. Specifically, it iteratively updates the pose so as to align the model's contours with the image ones.
As such, it may be sensitive to the target shape, object occlusion and background clutter.
Instead, we use dense pixel-wise correspondences, which are more robust to these disturbances and only need to be predicted once.

{\bf Optical flow estimation}, which provides dense 2D-to-2D correspondences between consecutive images~\cite{Jerome15a,Hu16,Ilg17,Hu17a,Sun18a,Teed20a}, is a building block of our framework. Rather than estimating the flow between two consecutive video frames, as commonly done by optical flow methods, we establish dense 2D correspondences between offline-generated synthetic exemplars and the real input image.
This is motivated by our observation that establishing dense correspondences between an image pair depends more strongly on the local differences between these images, rather than the images themselves, making this strategy more robust to a domain change. This is evidenced by the fact that our network trained only on synthetic data remains effective when applied to real images.


{\bf Domain adaptation} constitutes the standard approach to bridging the gap between different domains. However, most domain adaptation methods assume the availability of considerable amounts of data from the target domain, albeit unsupervised, to facilitate the adaptation~\cite{Tang20,Pan20,Gu20,Cui20}. Here, by contrast, we focus on the scenario where capturing such data is difficult. As such, domain generalization, which aims to learn models that generalize to unseen domains~\cite{Gulrajani21a}, seems more appropriate for solving our task. However, existing methods typically assume that multiple source domains are available for training, which is not fulfilled in our case. Although one can generate many different domains by augmentation techniques~\cite{Zhou21a,Wang21e,Xu21a}, we observed this strategy to only yield rough 6D pose estimates in the test domain. Therefore, we use this approach to obtain our initial pose, which we refine with our method.


\section{Approach}
\label{sec:approach}

\begin{figure*}[t]
    \begin{center}
    %
    \addtolength{\tabcolsep}{-5pt}  
    \begin{tabular}{cccc}
    \includegraphics[width=0.99\linewidth]{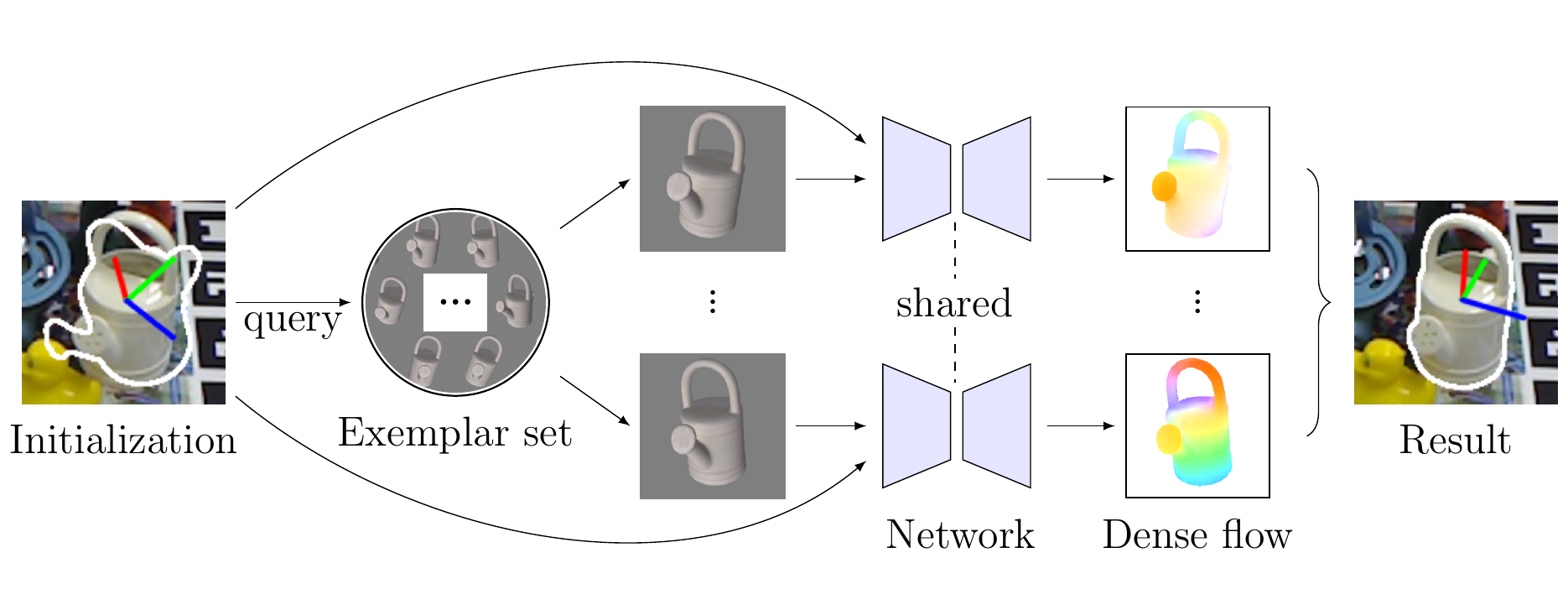}
    \end{tabular}
    \addtolength{\tabcolsep}{5pt}  
    \end{center}
    \vspace{-10mm}
    \caption{{\bf Overview of our framework.} We first obtain an initial pose of the target using a pose network trained only on synthetic data. We then retrieve the $N$ closest exemplars from the offline-rendered exemplar set and estimate their dense 2D displacement fields with respect to the target. Finally, we combine all these flow results into a set of 3D-to-2D correspondences to obtain a robust final pose estimate.}
    \label{fig:arch}
\end{figure*}

We aim to estimate the 6D pose of a known rigid object from a single RGB image in a data-limited scenario, that is, with little or even no access to real images during training. To this end, we use a  two-step strategy that first estimates a rough initial pose and then refines it. Where we differ from other methods is in our approach to refinement. Instead of using the usual iterative strategy, we introduce a non-iterative one that relies on an optical flow technique to estimate 2D-to-2D correspondences between an image rendered using the object's 3D model in the estimated pose and the target image. The required pose correction between the rough estimate and the correct one can then be computed algebraically using a PnP algorithm. Fig.~\ref{fig:arch} depicts our complete framework. 


\subsection{Data-Limited Pose Initialization}

Most pose refinement methods~\cite{Li18a,Zakharov19a,Yann20a} assume that rough pose estimates are provided by another approach trained on a combination of real and synthetic data~\cite{Xiang18b}, often augmented in some manner~\cite{Peng19a,Hu20a,Hu21a}. In our data-limited scenarios, real images may not be available, and we have to rely on synthetic images alone to train the initial pose estimation network. 

We will show in the results section that it requires very substantial augmentations for methods trained on synthetic data alone to generalize to real data, and that they only do so with a low precision. In practice, this is what we use to obtain our initial poses. 



\subsection{From Optical Flow to Pose Refinement}
\label{sec:flow_to_refine}


Given an imprecise estimate of the initial pose, we seek to refine it. To this end, instead of directly regressing a 6D pose correction from an image rendered using the pose estimate, we train a network to output a dense 2D-to-2D correspondence field between the rendered image and the target one, that is, to estimate optical flow~\cite{Teed20a}. From these dense 2D correspondences, we can then algebraically compute the 6D pose correction using a PnP algorithm. In the results section, we will show that this approach generalizes reliably to real images even when trained only on synthetic ones.


\begin{figure}[t]
    \begin{center}
    %
    \addtolength{\tabcolsep}{-5pt}  
    \includegraphics[width=0.99\linewidth]{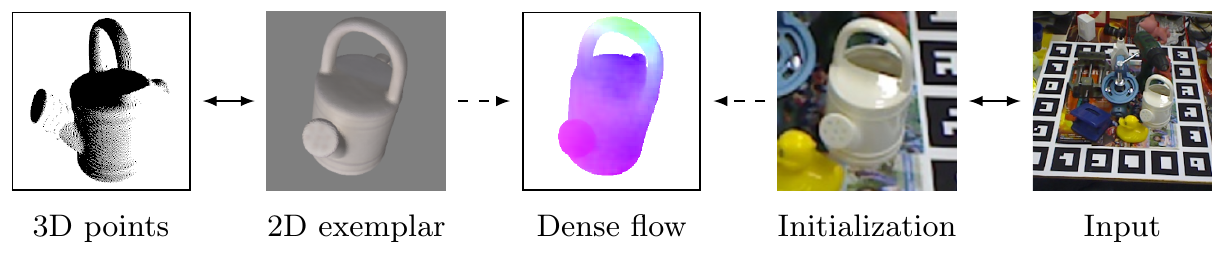}
    \addtolength{\tabcolsep}{5pt}  
    \end{center}
    \vspace{-6mm}
    \caption{{\bf Establishing 3D-to-2D correspondences from exemplars.} After retrieving an exemplar based on the initial pose, we estimate the 2D-to-2D correspondences between the exemplar and the input image within their respective region of interest. This implicitly generates a set of 3D-to-2D correspondences.
    }
    \label{fig:flow_examplar}
\end{figure}

More formally, let $\bI^t$ be the image of the target object and let $\bI^r$ be the one rendered using the rough pose estimate. We train a network to predict the 2D flow image $\bF^{r \rightarrow t}$ such that
\begin{align}
\forall i \in {\mathcal M}, \quad \bu_i^t  = \bu_{i}^r + \f_{i}^{r \rightarrow t} \; ,
\end{align}
where $\mathcal M$ contains the indices of the pixels in  $\bI^r$ for which a corresponding pixel in  $\bI^t$ exists, $\bu_i^t$ and $ \bu_{i}^r $ denote the pixel locations of matching points in both images, and 
$\f_{i}^{r \rightarrow t} $ is the corresponding 2D flow vector. 

Because $\bI^r$ has been rendered using a known 6D pose, the 2D image locations $\bu^r$ are in known correspondence with 3D object points ${\bf p}$. Specifically, the 3D point ${\bf p}_i$ corresponding to the 2D location $\bu_i^r$ can be obtained by intersecting the camera ray passing through $\bu_i^r$ and the 3D mesh model transformed by the initial 6D pose~\cite{Kato18}, as shown in the left of Fig.~\ref{fig:flow_examplar}. For each such correspondence, which we denote as $\{{\bf p}_i \leftrightarrow {\bf u}_{i}^r\}$, we have
\begin{equation}
    \begin{aligned}
    \lambda_i
    \begin{bmatrix}
    {\bf u}_i^r \\
    1 \\
    \end{bmatrix}
    =\bK(\bR{\bf p}_i+{\bf t}),
    \end{aligned}
    \label{eq:perspective}
\end{equation}
where $\lambda_i$ is a scale factor encoding depth, $\bK$ is the matrix of camera intrinsic parameters, and $\bR$ and ${\bf t}$ are the rotation matrix and translation vector representing the 6D pose. 

\begin{figure}[t]
    \begin{center}
    %
    \addtolength{\tabcolsep}{-5pt}  
    \begin{tabular}{c@{\hskip 0.2em}c@{\hskip 0.2em}c@{\hskip 0.2em}c@{\hskip 0.2em}c}
    %
    \includegraphics[width=0.18\linewidth]{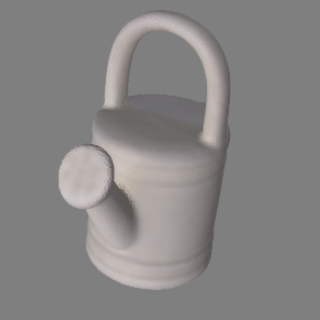} &
    \includegraphics[width=0.18\linewidth]{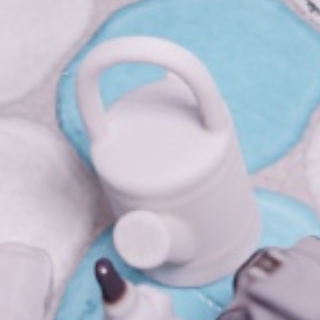}  &
     \includegraphics[width=0.18\linewidth]{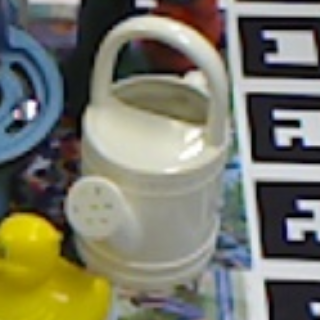}  &
    {\setlength{\fboxsep}{-0.5pt}\setlength{\fboxrule}{0.5pt}\fbox{\includegraphics[width=0.18\linewidth]{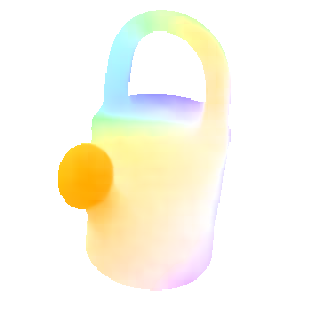}}}  &
     {\setlength{\fboxsep}{-0.5pt}\setlength{\fboxrule}{0.5pt}\fbox{\includegraphics[width=0.18\linewidth]{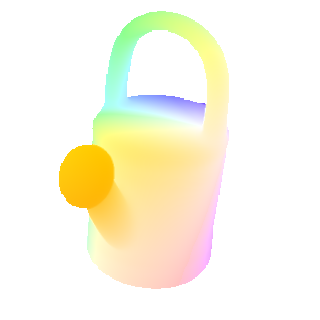}}}  \\
    {\small Exemplar} & {\small PBR input} & {\small Real input} & {\small Prediction} & {\small Ground truth} \\
    \end{tabular}
    \addtolength{\tabcolsep}{5pt}  
    \end{center}
    \vspace{-6mm}
    \caption{{\bf Estimating dense 2D-to-2D correspondences across domains.} We train a flow network to model the differences of images rendered using OpenGL (Exemplar) and using a PBR (Physically-based rendering) technique, respectively. Although our network accesses no real data during training, it generalizes well to estimating the flow between the exemplar and a real input image, as shown in the last two subfigures.
    }
    \label{fig:flow_demo}
\end{figure}

To simplify the discussion, let us for now assume that the intrinsic matrix $\bK$  used to render $\bI^r$ is the same as that of the real camera, which is assumed to be known by most 6D pose estimation methods. We will discuss the more general case in Section~\ref{sec:small_obj}.
Under this assumption, the flow vectors predicted for an input image provide us with 2D-to-3D correspondences between the input image and the 3D model. That is, for two image locations $(\bu_i^t, \bu_{i}^r)$ deemed to be in correspondence according to the optical flow, we have
\begin{equation}
\{{\bf p}_i \leftrightarrow {\bf u}_{i}^r\} \Leftrightarrow \{{\bf p}_i \leftrightarrow {\bf u}_{i}^t \}.
\label{eq:basic_flow}
\end{equation}
Given enough such 3D-to-2D correspondences, the 6D pose in the input image can be obtained algebraically using a PnP algorithm~\cite{Lepetit09}. In other words, we transform the pose refinement problem as a 2D optical flow one, and the 3D-to-2D correspondence errors will depend only on the 2D flow field ${\bf f}^{r \rightarrow t}$. 
Fig.~\ref{fig:flow_demo} shows an example of dense correspondences between the synthetic and real domains.


\subsection{Exemplar-Based Flow Aggregation}

The above-mentioned flow-based strategy suffers from the fact that it relies on an expensive rendering procedure, which slows down both training and testing. To address this, we use exemplars rendered offline. Instead of synthesizing the image from the initial pose directly, which requires on-the-fly rendering, we then retrieve the exemplar with 6D pose nearest to the initial pose estimate and compute the 2D displacements between this exemplar and the input image.

\begin{figure}[t]
    \begin{center}
    %
    \includegraphics[width=0.8\linewidth]{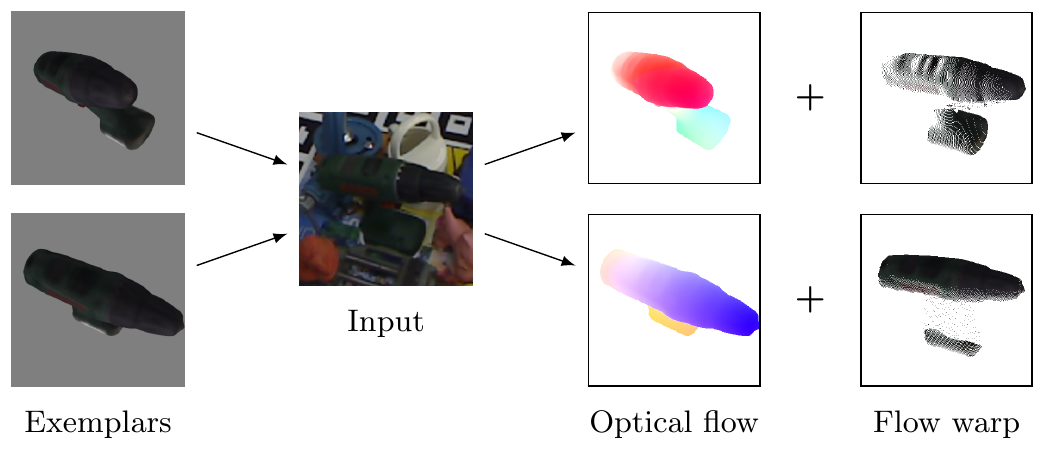}

    \addtolength{\tabcolsep}{5pt}  
    \end{center}
    \vspace{-6mm}
    \caption{{\bf Multi-view flow aggregation.} The multiple exemplars provide 3D-to-2D correspondences that are complementary since they are rendered from different viewpoints. These correspondences are then combined to make the final pose estimates more robust.
    }
    \label{fig:flow_aggregation}
\end{figure}

The resulting speed increase comes at the cost of a slight accuracy loss. However, it is compensated by the fact that this approach enables to exploit {\it multiple} rendered views, while only needing a single input image.
That is, we do not use a single exemplar but multiple ones rendered from different viewpoints to make our pose refinement more robust. During inference, we use the initial pose to find the $N$ closest exemplars and combine their optical flow. In short, instead of having one set of 3D-to-2D correspondences, we now have $N$ such sets, which we write as
\begin{equation}
    \{{\bf p}_{k,i} \leftrightarrow {\bf u}_{k,i}^t\} \quad 1\leq i \leq n_k \;, \;\;k \in \{1,\dots,N\}\;,
    \label{eq:flow_aggregation}
\end{equation}
where $n_k$ is the number of correspondences found for exemplar $k$. Because the exemplars may depict significantly different viewpoints, this allows us to aggregate more information and adds both robustness and accuracy, as depicted by Fig.~\ref{fig:flow_aggregation}. Finally, we use a RANSAC-based PnP algorithm~\cite{Lepetit09} to derive the final pose based on these complementary correspondences. 
 

\subsection{Dealing with Small Objects}
\label{sec:small_obj}

In practice, even when using multiple exemplars, the approach described above may suffer from the fact that estimating the optical flow of small objects is challenging. To tackle this, inspired by other refinement methods~\cite{Li18a,Zakharov19a,Yann20a}, we work on image crops around the objects. Specifically, because we know the ground-truth pose for the exemplars and have a rough pose estimate for the input image, we can define 2D transformation matrices ${\bf M}_r$ and ${\bf M}_t$ that will map the object region in the exemplar and in the input image to a common size. We can then compute the flow between the resulting transformed images.


Formally, let $\tilde{\bf u}_{i}^r = {\bf M}_r{\bf u}_{i}^r$ be an exemplar 2D image location after transformation. Furthermore, accounting for the fact that the intrinsic camera matrices used to render the exemplars and acquire the input image may differ, let $\tilde{\bf u}_{i}^t = {\bf M}_t{\bf K}_r{\bf K}_t^{-1}{\bf u}_i^t$ be an input image location after transformation, where ${\bf K}_r$ is the intrinsic matrix used for the exemplars and ${\bf K}_t$ the one corresponding to the input image. We then estimate a flow field $\{\tilde\f_{i}^{r \rightarrow t} = \tilde{\bf u}_{i}^t - \tilde{\bf u}_{i}^r\}$ using the flow network. For two transformed image locations $(\tilde{\bu}_i^t, \tilde{\bu}_{i}^r)$ found to be in correspondence, following the discussion in Section~\ref{sec:flow_to_refine}, we can establish 3D-to-2D correspondences in the transformed image as
\begin{equation}
    \{{\bf p}_i \leftrightarrow \tilde{\bf u}_{i}^r\} \Leftrightarrow \{{\bf p}_i \leftrightarrow \tilde{\bf u}_{i}^t \}\;.
    \label{eq:flow_examplar}
\end{equation}
We depict this procedure in Fig.~\ref{fig:flow_examplar}. We can then recover the corresponding ${\bf u}_{i}^r$ in the original input image by applying the inverse transformation, which then lets us combine the correspondences from multiple exemplars.

\comment{
we only work on patches containing the targets similar to , as we know the initial pose which already can give us the rough position of targets on the image. This strategy, however, will make the image-level correspondence relation in Eq.~\ref{eq:basic_flow} not directly applicable.

We first make our training of the 2D flow network independent of different real camera matrix. For all the rendering processes, we use a single internal camera matrix ${\bf K_1}$, which may be different from that of the input ${\bf K_2}$. We convert the input to have the same camera matrix by a 2D transformation ${\bf K_1}{\bf K_2^{-1}}$ to make them consistent before the processing. 
Given the ground truth pose of the exemplar and the rough initial pose of the input, we can transform the targets to a similar size level and center-aligned by 2D transformation matrix ${\bf M_1}$ and ${\bf M_2}$, according to a simple 2D projection of the 3D mesh, respectively. 
Then we will estimate the dense correspondence field ${\bf f}$ between these two patches. 
Following the discussion in Section~\ref{sec:flow_to_refine}, we can easily establish patch-level 3D-to-2D correspondences for the real input
\begin{equation}
    \{{\bf p}_i \leftrightarrow \tilde{\bf u}_{i}^r\} \Leftrightarrow \{{\bf p}_i \leftrightarrow \tilde{\bf u}_{i}^t \}
    \label{eq:flow_examplar}
\end{equation}
where $\tilde{\bf u}_{i}^r = {\bf M_1}{\bf u}_{i}^r$ and $\tilde{\bf u}_{i}^t = {\bf M_2}{\bf K_1}{\bf K_2^{-1}}{\bf u}_i^t$, and the patch-level flow field $\{\tilde\f_{i}^{r \rightarrow t} = \tilde{\bf u}_{i}^t - \tilde{\bf u}_{i}^r\}$ will be estimated by the flow network, which is much easier than working in the raw image scale. We depict this procedure in Fig.~\ref{fig:flow_examplar}. To combine multiple patch-level flow fields for flow aggregation, we will recover ${\bf u}_{i}^r$ and ${\bf u}_{i}^t$ first through the reverse transformations to make all 2D matching points at the same target image plane, making them combinable.
}
\subsection{Implementation Details}
\label{sec:impl_detail}

We use the WDR-Pose network~\cite{Hu21a} as our initialization network, and RAFT~\cite{Teed20a} as our 2D correspondence network. We first train WDR-Pose on the BOP synthetic data~\cite{Hodan18,hodan19a}, which contains multiple rendered objects and severe occlusions in each frame to simulate real images.
Before training the flow network, we generate a set of exemplars for each object type by offline rendering. To avoid computing a huge set of exemplars by densely sampling the 6D pose space, we fix the 3D translation and randomly sample a small set of 3D rotations. Specifically, we set the 3D translation to $(0,0,\bar{z})$, where $\bar{z}$ is approximately the average depth of the working range. In our experiments, we found that 10K exemplars for each object type yields a good accuracy. We generate our exemplar images using the method of~\cite{Kato18} with a fixed light direction pointing from the camera center to the object center.

To build image pairs to train the flow network, we pick one image from the exemplar set and the other from the BOP synthetic dataset. Specifically, we query the closest exemplar  in terms of 6D pose. To simulate the actual query process, we first add some pose jitter to the target instance. Specifically we add a random rotation angle within 20 degrees and a random translation leading to a reprojection offset smaller than 10 pixels. Note that this randomness only affects the query procedure; it does not affect the supervision of the flow network, which relies on the selected exemplar and the synthetic image.

In practice, to account for small objects, we first extract the object instances from the exemplar and target images using the transformation matrices ${\bf M}_r$ and ${\bf M}_t$ discussed above. We then resize the resulting object crops to $256\times 256$ and build the ground-truth flow according to Eq.~\ref{eq:flow_examplar}. We only supervise the flow of pixels located within the exemplar's object mask. Furthermore, during training, we also discard all pixels without explicit correspondence because of occlusions or because they fall outside the crops.

Finally, during training, we apply two main categories of data augmentation techniques. The first is noise (NS) augmentation. We add a random value between -25 to 25 to the pixels in each image channel. We then blur the resulting image with a random kernel size between 1 and 5. The second is color (HSV) augmentation. We convert the input image from RGB to HSV and add random jitter to each channel. Specifically, we add 20\%, 50\%, 50\% of the maximum value of each channel as the random noise to the H, S, and V channel, respectively. We then convert the image back to RGB.

\section{Experiments}
\label{sec:experiments}

In this section, we first compare our approach to the state of the art on standard datasets including LINEMOD (``LM'')~\cite{Hinterstoisser12b}, Occluded-LINEMOD (``OLM'')~\cite{Krull15} and YCB-V (``YCB'')~\cite{Xiang18b}. We then evaluate the influence of different components of our refinement network. We defer the evaluation of the initialization network trained only on synthetic data to Section~\ref{sec:init}. The source code is publicly available at \href{https://github.com/cvlab-epfl/perspective-flow-aggregation}{https://github.com/cvlab-epfl/perspective-flow-aggregation}.

\noindent{\bf Datasets and Experimental Settings.} LINEMOD comprises 13 sequences. Each one contains a single object annotated with the ground-truth pose. There are about 1.2K images for each sequence. We train our model only on the exemplars and the BOP synthetic dataset, and test it on 85\% of the real LINEMOD data as in~\cite{Rad17,Li18a}. We keep the remaining 15\% as supplementary real data for our ablation studies. Occluded-LINEMOD has 8 objects, which are a subset of the LINEMOD ones. It contains only 1214 images for testing, with multiple objects in the same image and severe occlusions, making it much more challenging. 

Most methods train their models for LINEMOD and Occluded-LINEMOD separately~\cite{Hu19a,Li19a}, sometimes even one model per object~\cite{Peng19a,Wanggu21a}, which yields better accuracy but is less flexible and does not scale well. As these two datasets share the same 3D meshes, we train a single model for all 13 objects and test it on LINEMOD and Occluded-LINEMOD without retraining. When testing on Occluded-LINEMOD, we only report the accuracy of the corresponding 8 objects. We show that our model still outperforms most methods despite this generalization that makes it more flexible.

YCB-V is a more recent dataset that contains 92 video sequences and about 130K real images depicting 21 objects in cluttered scenes with many occlusions and complex lighting conditions. As for LINEMOD, unless stated otherwise, we train our model only on the exemplars and the BOP synthetic dataset and test on the real data.

\noindent{\bf Evaluation metrics.}
We compute the 3D error as the average distance between 3D points on the object surface transformed by the predicted pose and by the ground-truth one. We then report the standard ADD-0.1d metric~\cite{Xiang18b}, that is, the percentage of samples whose 3D error is below 10\% of the object diameter. For more detailed comparisons, we use ADD-0.5d, which uses a larger threshold of 50\%. Furthermore, to compare with other methods on YCB-V, we also report the AUC metric as in~\cite{Peng19a,Yann20a,Wanggu21a}, which varies the threshold with a maximum of 10cm and accumulates the area under the accuracy curve. For symmetric objects, the 3D error is taken to be the distance of each 3D point to its nearest model point after pose transformation.



\begin{table}[t]
	\centering
	\caption{{\bf Comparing against the state of the art.} Our method trained without accessing any real images (+0) performs on par with most methods that use \emph{all} real data (hundreds of images per object for LM and OLM, and thousands for YCB). After accessing only 20 real images per object (+20), our method yields the best results.
    }
	\scalebox{0.83}{
	\begin{small}
	\begin{tabular}{lccccccccc}
	\toprule
    Data & Metrics & \multicolumn{1}{c}{PoseCNN} & \multicolumn{1}{c}{SegDriven} & \multicolumn{1}{c}{PVNet} & \multicolumn{1}{c}{GDR-Net} & \multicolumn{1}{c}{DeepIM} & \multicolumn{1}{c}{CosyPose} & \multicolumn{1}{c}{\bf Ours (+0)} & \multicolumn{1}{c}{\bf Ours (+20)} \\
	\midrule
	LM& ADD-0.1d  & 62.7 & -& 86.3& 93.7 & 88.6 & - & 84.5 & {\bf 94.4}\\
	OLM & ADD-0.1d &24.9&27.0&40.8&62.2& 55.5 & -  & 48.2 & {\bf 64.1}\\
	\cmidrule{2-10}
	\multirow{2}{*}{YCB} & ADD-0.1d &21.3&39.0 & - & 60.1& - & -  & 56.4 & {\bf 62.8}\\
	& AUC & 61.3 &-& 73.4 & 84.4 & 81.9 & 84.5 & 76.8 & {\bf 84.9}\\
	\bottomrule
	\end{tabular}
	\end{small}
	}
    \label{tab:stoa}
	\vspace{-3mm}
\end{table} 

\subsection{Comparison with the State of the Art}

\begin{figure*}[t]
    \begin{center}
    %
    \addtolength{\tabcolsep}{-5pt}  
    \begin{tabular}{c@{\hskip 0.2em}c@{\hskip 0.2em}c@{\hskip 0.2em}c@{\hskip 0.2em}c}
    \includegraphics[width=0.18\linewidth]{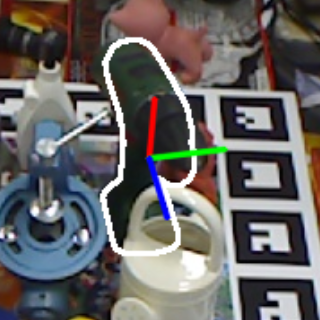} &
    \includegraphics[width=0.18\linewidth]{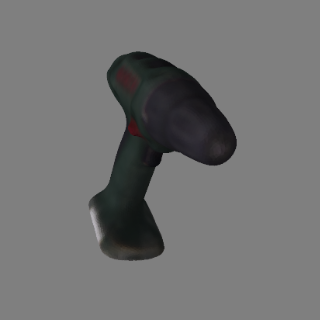} &
    {\setlength{\fboxsep}{-0.5pt}\setlength{\fboxrule}{0.5pt}\fbox{\includegraphics[width=0.18\linewidth]{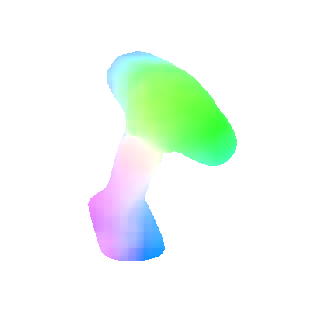}}} &
    {\setlength{\fboxsep}{-0.5pt}\setlength{\fboxrule}{0.5pt}\fbox{\includegraphics[width=0.18\linewidth]{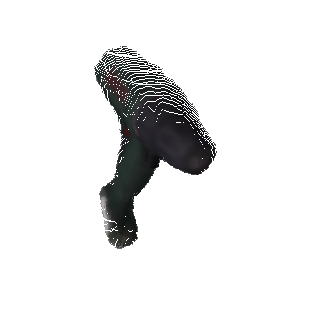}}} &
    \includegraphics[width=0.18\linewidth]{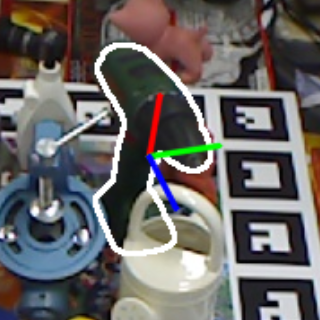}\\
    \includegraphics[width=0.18\linewidth]{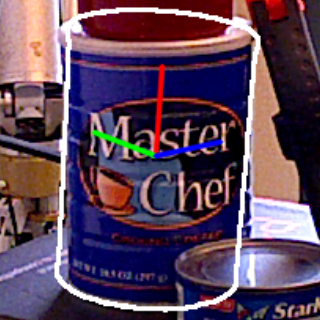} &
    \includegraphics[width=0.18\linewidth]{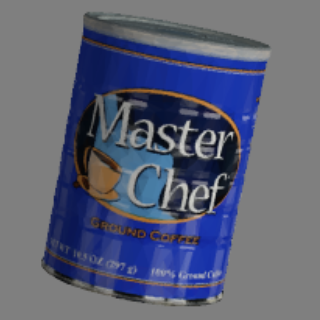} &
    {\setlength{\fboxsep}{-0.5pt}\setlength{\fboxrule}{0.5pt}\fbox{\includegraphics[width=0.18\linewidth]{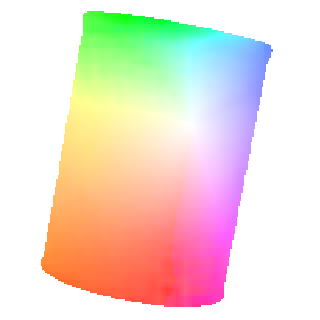}}} &
    {\setlength{\fboxsep}{-0.5pt}\setlength{\fboxrule}{0.5pt}\fbox{\includegraphics[width=0.18\linewidth]{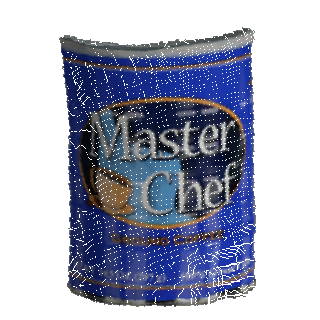}}} &
    \includegraphics[width=0.18\linewidth]{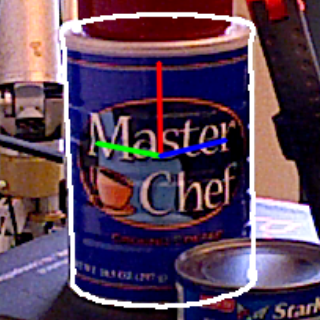}\\
    {\scriptsize Initializations} & {\scriptsize Exemplars} & {\scriptsize Predicted flow} & {\scriptsize Forward warp} & {\scriptsize Refinements} \\
    \end{tabular}
    \addtolength{\tabcolsep}{5pt}  
    \end{center}
    \vspace{-7mm}
    \caption{{\bf Visualization of the results.} 
    Although the predicted flows contain some errors (e.g., the flow at the bottom of the drill, which is  occluded), aggregating multiple flows and using a RANSAC-based PnP make the final pose estimation robust. Here we show results obtained with one exemplar and by training purely on synthetic data.
    }
    \label{fig:visualization}
\end{figure*}
\begin{figure}[t]
    \begin{center}
    %
    \addtolength{\tabcolsep}{-5pt}  
    \begin{tabular}{c@{\hskip 0.2em}c@{\hskip 0.2em}c@{\hskip 0.2em}c}
    \includegraphics[width=0.19\linewidth]{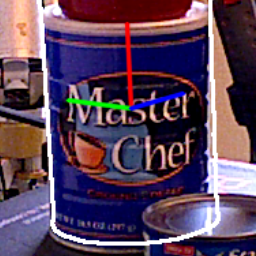}  &
    \includegraphics[width=0.19\linewidth]{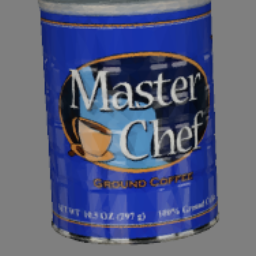} &
    \includegraphics[width=0.19\linewidth]{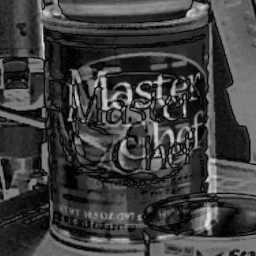} &
    \includegraphics[width=0.19\linewidth]{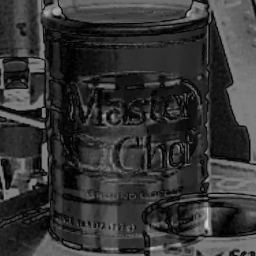} \\
    {\scriptsize Annotation} & {\scriptsize Reprojection} & {\scriptsize Difference} & {\scriptsize Our result} \\
    \end{tabular}
    \addtolength{\tabcolsep}{5pt}  
    \end{center}
    \vspace{-7mm}
    \caption{{\bf An example of inaccurate annotation in YCB-V.} The last two images show the difference between the input and the reprojection image rendered from the corresponding pose. Our predicted pose aligns the object more accurately here.
    }
    \label{fig:gt_eval}
\end{figure}

We now compare our method to the state-of-the-art ones, PoseCNN~\cite{Xiang18b}, SegDriven~\cite{Hu19a}, PVNet~\cite{Peng19a}, GDR-Net~\cite{Wanggu21a}, DeepIM~\cite{Li18a}, and CosyPose~\cite{Yann20a}, where DeepIM and CosyPose are two refinement methods based on an iterative strategy. 
We train our initialization network WDR-Pose only on synthetic data and use its predictions as initial poses. To train the optical flow network, we generate 10K exemplars for each object and use the $N=4$ closest exemplars during inference.
As shown in Table~\ref{tab:stoa}, even without accessing any real images during training, our method already outperforms most of the baselines, which all use real training data, and performs on par with the most recent ones. Fig.~\ref{fig:visualization} depicts some qualitative results. Note that YCB-V contains some inaccurate pose annotations, and, as shown in Fig.~\ref{fig:gt_eval}, we sometimes predict more accurate poses than the annotations, even when training without accessing any real data.

In Table~\ref{tab:stoa}, we also report the results we obtain by adding 20 real images to the synthetic ones during the training of our refinement network. 
In this case, we train each model by mixing the BOP synthetic data with the real images, balancing their ratio to be 0.5:0.5 by dynamic sampling during training. 
Note that we still use the same pose initializations trained only on synthetic images.
With only 20 real images, our method outperforms all the baselines by a significant margin, even though they all use {\it all} the real images during training.




\begin{table}[t]
	\centering
	\caption{{\bf Comparing different refinement frameworks on OLM.} We train the models with different numbers of additional real images. With as few as 20 real images, our model achieves even higher accuracy than the baselines with more than 100 real images.
    }
	\scalebox{0.83}{
	\begin{small}
	\begin{tabular}{l@{\hskip 2em}c@{\hskip 1em}c@{\hskip 1em}c@{\hskip 1em}c@{\hskip 1em}c}
	\toprule
     & 0 & 10 & 20 & 90 & 180 \\
	\midrule
	DeepIM & 41.1 & 45.6 & 48.2 & 58.1 & 61.4 \\
	CosyPose & 42.4 & 46.8 & 48.9 & 58.8 & 61.9 \\
	\bf{Ours} & {\bf 48.2} & {\bf 59.5} & {\bf 64.1} & {\bf 64.9} & {\bf 65.3}\\
	\bottomrule
	\end{tabular}
	\end{small}
	}
    \label{tab:real_data}
	\vspace{-3mm}
\end{table}

Most existing refinement methods, including DeepIM and CosyPose, employ pose initializations trained using all real images, which is impractical in our data-limited scenario. To have a fair comparison of our refinement method with them, we use the same synthetic-only pose initializations for them as for our approach. We then train the refinement networks according to their open-source official code, based on synthetic data only. Furthermore, we also evaluate them when trained with different numbers of additional real images. Note that, while CosyPose can use multiple views as input, we only evaluate it in the monocular case to make it comparable with the other methods.
In Table~\ref{tab:real_data}, we report the ADD-0.1d on the challenging Occluded-LINEMOD dataset. As expected, using more real images yields more accurate pose estimates for all methods. However, with as few as 20 real images, our model achieves even higher accuracy than the baselines with more than 100 real images. Since all methods use the same initial poses, giving an accuracy of 37.9\%, as shown in Table~\ref{tab:eval_of_init}, this means that DeepIM and CosyPose can only increase the initialization accuracy by 3-4\% when not accessing any real image data. By contrast, our method increases accuracy by over 10\% in this case, demonstrating the robustness of our method to the domain gap in cluttered scenarios. 

\subsection{Ablation Studies}
\label{sec:refine}
Let us now analyze more thoroughly the exemplar-based flow aggregation in our pose refinement framework. To this end, we conduct more ablation studies on the standard LINEMOD dataset. We train our refinement model only on synthetic data~\cite{hodan19a}, and report ADD-0.1d accuracies on real test data.

\noindent{\bf Flow Aggregation.}
We first evaluate our flow aggregation strategy given different pose initializations. We use three initialization sets with varying levels of accuracy, corresponding to the results from the initialization network under NS, HSV, and NS+HSV augmentations, respectively. Furthermore, we evaluate the accuracy using different numbers of retrieved exemplars, also reporting the corresponding running speed on a typical workstation with a 3.5G CPU and an NVIDIA V100 GPU.


\begin{table}[t]
	\centering
	\caption{{\bf Pose refinement with different initializations on LM.} With only one exemplar ($N$=1), our refinement framework already yields a significant improvement over the initialization. More exemplars make it more accurate.
	}
	\scalebox{0.83}{
	\begin{small}
	\begin{tabular}{l@{\hskip 1em}c@{\hskip 2em}c@{\hskip 1em}c@{\hskip 1em}c@{\hskip 1em}c@{\hskip 1em}}
	\toprule
    \multicolumn{2}{c@{\hskip 2em}}{Initialization} & {$N$=1} & {$N$=2} & {$N$=4} & {$N$=8} \\
	\midrule
	NS & 54.1 & 82.0 & 82.7& {\bf 84.3} & 84.1\\
	HSV & 52.7 & 81.1 & 81.2& 83.3& {\bf 83.9}\\
	NS+HSV & 60.2 & 82.0 & 83.4 & 84.5 & {\bf 84.9}\\
	\midrule
	\multicolumn{2}{c}{FPS} & $\sim$32 & $\sim$25 & $\sim$20 & $\sim$14\\
	\bottomrule
	\label{tab:flow_aggregation}
	\end{tabular}
	\end{small}
	}
	\vspace{-3mm}
\end{table} 

As shown in Table~\ref{tab:flow_aggregation}, the refinement network improves the accuracy of the initial pose significantly even with only one exemplar. More exemplars boost it  further, thanks to the complementarity of their different viewpoints. 
Interestingly, although the different pose initializations have very different pose accuracies, they all reach a similar accuracy after our pose refinement, demonstrating the robustness of our refinement network to different initial poses.
As the exemplars can be processed in parallel, the running time with 4 exemplars is only about 1.6 slower than that with a single exemplar. This slight speed decrease is related to the throughput of the GPU and could be optimized further in principle. Nevertheless, our approach is still more than 3 times faster than the iterative DeepIM~\cite{Li18a} method, which runs at only about 6 FPS using 4 iterations. 
Since the version with 8 exemplars yields only a small improvement over the one with 4 exemplars, we use $N$=4 in the previous experiments. Furthermore, we use the results of NS+HSV for pose initialization.

\noindent{\bf Exemplar Set.}
While using an exemplar set eliminates the need for online rendering, the accuracy of our approach depends on its granularity, leading to a tradeoff between accuracy and IO storage/speed. We therefore evaluate the performance of our approach with varying numbers of exemplars during inference. 
To better understand the query process, we also report the numbers just after the query but before the refinement, denoted as ``Before Ref.''. 

Table~\ref{tab:exemplar_set_size} shows that larger exemplar sets yield more accurate queries before the refinement, leading to more accurate pose refinement results. Note that because we have a discrete set of exemplars, the ADD-0.1d scores before refinement are lower than those obtained by on-the-fly rendering from the initial pose, which reaches 60.2\%. 
While fewer exemplars in the set translates to lower accuracy before refinement, the accuracy after refinement saturates beyond 10K exemplars, reaching a similar performance to online rendering. We therefore use 10K exemplars for each object in the previous experiments. This only requires about 200MB of disk space for storing the exemplar set for each object. 
We also report the timings of image preparation for each setting. Although there is a powerful GPU for the online rendering, our offline exemplar retrieval is much faster.



\begin{table}[t]
	\centering
	\caption{{\bf Effect of exemplar sets of different sizes on LM.} An exemplar set with less than 5K exemplars suffers from the large distance between the nearest exemplars. By contrast, using more than 10K exemplars does not bring much improvement.
	}
	\scalebox{0.83}{
	\begin{small}
	\begin{tabular}{l@{\hskip 1em}c@{\hskip 1em}c@{\hskip 1em}c@{\hskip 1em}c@{\hskip 1em}c@{\hskip 1em}c}
	\toprule
		& 2.5K & 5K & 10K & 20K & 40K & Online\\
	\midrule
	Before Ref. & 55.2 & 57.0 & 58.0 & 58.9 & 59.9 & {\bf 60.2}\\
	After Ref. & 81.9 & 83.2 & 84.5 & {\bf 85.0} & 84.8 & 84.9\\
	\midrule
	Image preparation & 16ms & 19ms & 29ms & 52ms & 81ms & 184ms \\
	\bottomrule
	\label{tab:exemplar_set_size}
	\end{tabular}
	\end{small}
	}
	\vspace{-3mm}
\end{table} 

\subsection{Pose Initialization Network}
\label{sec:init}

We now evaluate our pose initialization network based on WDR-Pose~\cite{Hu21a}. Unlike in~\cite{Hu21a}, we train it only on the BOP synthetic data~\cite{Hodan18,hodan19a} and study the performance on real images. To fill the domain gap between the synthetic and real domains, we use simple data augmentation strategies during training. 

Specifically, we evaluate 3 groups of data augmentations. The first one consists of random shifts, scales, and rotations within a range of (-50px, 50px), (0.9, 1.1), and (-45$^\circ$, 45$^\circ$), respectively. We refer to this as SSR augmentations. The second group incudes random noise and smoothness, and corresponds to the NS augmentations discussed in Section~\ref{sec:refine}. The final group performs color augmentations, and corresponds to the HSV augmentations presented in Section~\ref{sec:refine}.



\begin{table}[t]
	\centering
	\caption{{\bf Data augmentation in pose initialization.} We study both geometric (SSR) and non-geometric (NS, HSV) augmentation strategies for pose initialization trained only on synthetic data but tested on real images.
    }
	\scalebox{0.83}{
	\begin{small}
	\begin{tabular}{c@{\hskip 1em}c@{\hskip 2em}c@{\hskip 1em}c@{\hskip 1em}c@{\hskip 1em}c@{\hskip 0.5em}c}
	\toprule
    Data & Metrics & {No} & {SSR} & {NS} & {HSV} & {NS+HSV} \\
	\midrule
	\multirow{2}{*}{LM} & ADD-0.1d & 48.9 & 39.0 & 56.6 & 58.1 & {\bf 60.2}\\
	& ADD-0.5d & 93.2 & 80.7 & 98.8 & 97.0 & {\bf 98.9}\\
	\cmidrule{2-7}
	\multirow{2}{*}{OLM} & ADD-0.1d & 28.6 & 20.9 & 36.0 & 34.4 & {\bf 37.9}\\
	& ADD-0.5d & 74.8 & 69.2 & 83.0 & 81.3 & {\bf 86.1}\\
	\cmidrule{2-7}
	\multirow{2}{*}{YCB} & ADD-0.1d & 0.1 & 0 & 17.0 & 7.4 & {\bf 27.5}\\
	& ADD-0.5d & 2.9 & 0 & 59.0 & 36.6 & {\bf 72.3}\\
	\bottomrule
	\end{tabular}
	\end{small}
	}
    \label{tab:eval_of_init}
	\vspace{-3mm}
\end{table}

Table~\ref{tab:eval_of_init} summarizes the results of the model trained with these different data augmentations on LINEMOD, Occluded-LINEMOD, and YCB-V. We report the accuracy in terms of both ADD-0.1d and ADD-0.5d. In short, training on synthetic data without data augmentation (``No'') yields poor performance on the real test data, with an accuracy of almost zero in both metrics on the YCB-V dataset. Interestingly, although the NS and HSV augmentations can considerably increase the performance, the SSR augmentations degrade it consistently on all three datasets. We believe this to be due to the geometric nature of the SSR augmentations.
More precisely, after shifting, scaling, or rotating the input image, the resulting inputs do not truly correspond to the original 6D poses, which inevitably introduces errors in the learning process. However, the NS and HSV augmentations do not suffer from this problem, as the ground-truth annotations before and after augmentation are the same. 

Note that, although the NS and HSV augmentations significantly outperform no augmentation, the accuracy remains rather low in terms of ADD-0.1d. However, the ADD-0.5d numbers evidence that most of the predictions have an error of less than 50\% of the diameter of the object. This indicates that the resulting rough initialization can indeed serve as a good starting point for our pose refinements, as demonstrated before.


\section{Conclusion}
\label{sec:conclusion}

We have introduced a simple non-iterative pose refinement strategy that can be trained only on synthetic data and yet still produce good results on real images. It relies on the intuition that, using data augmentation, one can obtain a rough initial pose from a network trained on synthetic images, and that this initialization can be refined by predicting dense 2D-to-2D correspondences between an image rendered in approximately the initial pose and the input image. Our experiments have demonstrated that our approach yields results on par with the state-of-the-art methods that were trained on real data, even when we don't use any real images, and outperforms these methods when we access as few as 20 images. In other words, our approach provides an effective and efficient strategy for data-limited 6D pose estimation. 
Nevertheless, our method remains a two-stage framework, which may limit its performance. In the future, we will therefore investigate the use of a differentiable component to replace RANSAC PnP and make our method end-to-end trainable.

\vspace{0.2em}
{\noindent \bf Acknowledgments.}
This work was supported by the Swiss Innovation Agency (Innosuisse). We would like to thank S\'ebastien Speierer and Wenzel Jakob in EPFL Realistic Graphics Lab (RGLab) for the helpful discussions on rendering.

%
%
\bibliographystyle{splncs04}
\bibliography{string,graphics,vision,learning,space}
\end{document}